\documentclass{bmvc2k}
\usepackage{booktabs}
\usepackage{multirow}
\usepackage{float}

\newcommand\blfootnote[1]{%
\begingroup
\renewcommand\thefootnote{}
\footnote{#1}%
\addtocounter{footnote}{-1}%
\endgroup
}

\title{DABNet: Depth-wise Asymmetric Bottleneck for Real-time Semantic Segmentation}

\addauthor{Gen Li}{ligen@skku.edu}{1}
\addauthor{Inyong Yun}{iyyun@skku.edu}{1}
\addauthor{Jonghyun Kim}{jhkim.ben@gmail.com}{1}
\addauthor{Joongkyu Kim\textsuperscript{$\dagger$}}{jkkim@skku.edu}{1}

\addinstitution{
 Department of Electronic, Electrical\\
 and Computer Engineering\\
 Sungkyunkwan University\\
 Suwon, South Korea
}
\runninghead{Li, \rm et al.}{DABNet for real-time semantic segmentation}

\begin{document}
\maketitle

\blfootnote{$\dagger$Corresponding author}

\begin{abstract}
As a pixel-level prediction task, semantic segmentation needs large computational cost with enormous parameters to obtain high performance.
Recently, due to the increasing demand for autonomous systems and robots, it is significant to make a trade-off between accuracy and inference speed.
In this paper, we propose a novel Depth-wise Asymmetric Bottleneck (DAB) module to address this dilemma, which efficiently adopts depth-wise asymmetric convolution and dilated convolution to build a bottleneck structure. 
Based on the DAB module, we design a Depth-wise Asymmetric Bottleneck Network (DABNet) especially for real-time semantic segmentation, which creates sufficient receptive field and densely utilizes the contextual information.
Experiments on Cityscapes and CamVid datasets demonstrate that the proposed DABNet achieves a balance between speed and precision.
Specifically, without any pretrained model and post-processing, it achieves 70.1\% Mean IoU on the Cityscapes test dataset with only 0.76 million parameters and a speed of 104 FPS on a single GTX 1080Ti card.
Code is available at \url{https://github.com/Reagan1311/DABNet}.
\end{abstract}

\section{Introduction}

\label{sec:intro}

Recently, autonomous systems and augmented reality devices have drawn widespread attention, which are the main application fields of semantic segmentation.
Such real-world applications not only demand competitive performance with low energy and memory but also have a strict requirement for inference speed.
Existing real-time semantic segmentation \cite{ENet,ESPNet, ContextNet,SkipNet-MobileNet} models have successfully sped up to real time, but meanwhile, they significantly sacrifice model accuracy.
Therefore, how to design an effective real-time semantic segmentation network with fast inference speed and small capacity becomes a challenging problem.

Most of the previous excellent work \cite{DeepLab, DeepLabv2, Deeplabv3, DilatedConv, Dilation8} has already proved the availability of dilated convolution which is able to create sizeable receptive field while maintaining the number of parameters.
Thus, many existing semantic segmentation models aiming at real time employ dilated convolution in their network.
Another method to effectively reduce the number of parameters is depth-wise separable convolution (ds-Conv). It computes cross-channel and spatial correlations independently, which is widely used in lightweight models \cite{MobileNet, MobileNetv2, Xception}.
However, simply replacing the standard convolution with ds-Conv results in large performance degradation, as ds-Conv dramatically reduces the parameters, and it often leads to a sub-optimal problem.
Therefore, here we hold the opinion that a well-designed combination of dilated convolution and ds-Conv is especially suitable for real-time semantic segmentation.

\begin{figure}[tbp]
\vspace{0.4cm}
\centering
\includegraphics[scale=0.19]{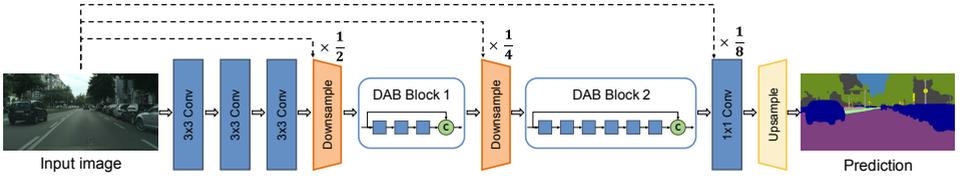}
\caption{Architecture of proposed Depth-wise Asymmetric Bottleneck Network. ``C'' means concatenation, dashed lines indicate average pooling operation.}
\label{Figure1}
\end{figure}

Based on the above observation, we propose a novel network architecture specially designated for real-time semantic segmentation, which is presented in Figure \ref{Figure1}.
More specifically, we design a depth-wise asymmetric bottleneck to extract dense feature under a shallow network, which has common advantages of both dilated convolution and depth-wise separable convolution.

Our main contributions could be summarized as:
\begin{itemize}
    \item We propose a bottleneck structure which incorporates depth-wise asymmetric convolution with dilated convolution, and it is termed as Depth-wise Asymmetric Bottleneck (DAB). This structure extracts local and contextual information jointly and dramatically reduces the parameters.
    \item We elaborately design DABNet based on DAB module,  it has much fewer parameters than the existing state-of-the-art real-time semantic segmentation methods while providing comparable accuracy and faster inference speed.
    \item We achieve remarkable results on both of Cityscapes \cite{Cityscapes} and CamVid \cite{CamVid} benchmarks without any context module, pretrained model, and post-processing scheme. Under an equivalent number of parameters, the proposed DABNet significantly outperforms existing semantic segmentation networks. It can run on high-resolution images ($512\times1024$) at 104 FPS on a single GTX 1080Ti card and 70.1\% Mean IoU on the Cityscapes test dataset with merely 0.76 M parameters.
\end{itemize}

\section{Related Work}
Recent real-time semantic segmentation models use multiple different techniques, such as dilated convolution and convolution factorization, to enlarge the receptive field and speed up networks.
In this section, we briefly introduce recent research progress and describe some approaches that are of great help in real-time semantic segmentation.

\noindent\textbf{Real-time semantic segmentation.}
Real-time semantic segmentation network requires finding a trade-off between high-quality prediction and high-inference speed.
ENet \cite{ENet} is the first network to be designed in real time, it trims a great number of convolution filters to reduce computation.
ICNet \cite{ICNet} proposes an image cascade network that incorporates multi-resolution branches.
ERFNet \cite{ERFNet} uses residual connections and factorized convolutions to remain efficient while retaining remarkable accuracy.
More recently, ESPNet \cite{ESPNet} introduces an efficient spatial pyramid (ESP), which brings great improvement in both speed and performance.
BiSeNet \cite{BiSeNet} proposes two paths to combine spatial information and context information.
These networks successfully made a trade-off between speed and performance, but there is still sufficient space for further improvement.

\noindent\textbf{Dilated convolution.}
Dilated convolution \cite{DilatedConv} inserts zeros between each pixel in a standard convolution, which leads to a large effective receptive field without increasing parameters, hence it is generally used in semantic segmentation models.
In DeepLab series \cite{DeepLabv2,Deeplabv3,Deeplabv3+}, an atrous spatial pyramid pooling (ASPP) module is introduced which employs multiple parallel filters with different dilation rates to collect multi-scale information. 
DenseASPP \cite{DenseASPP} concatenates a set of dilated convolution layers to generate dense multi-scale feature representation.
Most of the state-of-the-art networks in semantic segmentation exploit dilated convolution, which proves its effectiveness in pixel-level prediction task.

\noindent\textbf{Convolution Factorization.}
Convolution factorization divides a standard convolution operation into several steps to reduce the computational cost and memory,  which is extensively adopted in lightweight CNN models.
Inception \cite{Inceptionv2,Inceptionv3,Inceptionv4} employ several small-sized convolutions to replace the convolution with large kernel size while maintaining the size of the receptive field.
Xception \cite{Xception} and MobileNet \cite{MobileNet} use the depth-wise separable convolution to reduce the amount of computation with only a slight drop in performance.
MobileNetV2 \cite{MobileNetv2} proposes an inverted residual block and linear bottlenecks to further improve the performance.
ShuffleNet \cite{ShuffleNet} applies the point-wise group convolution with channel shuffle operation to enable information communication between different groups of channels.

\begin{figure}[tbp]
\centering
\includegraphics[scale=0.4]{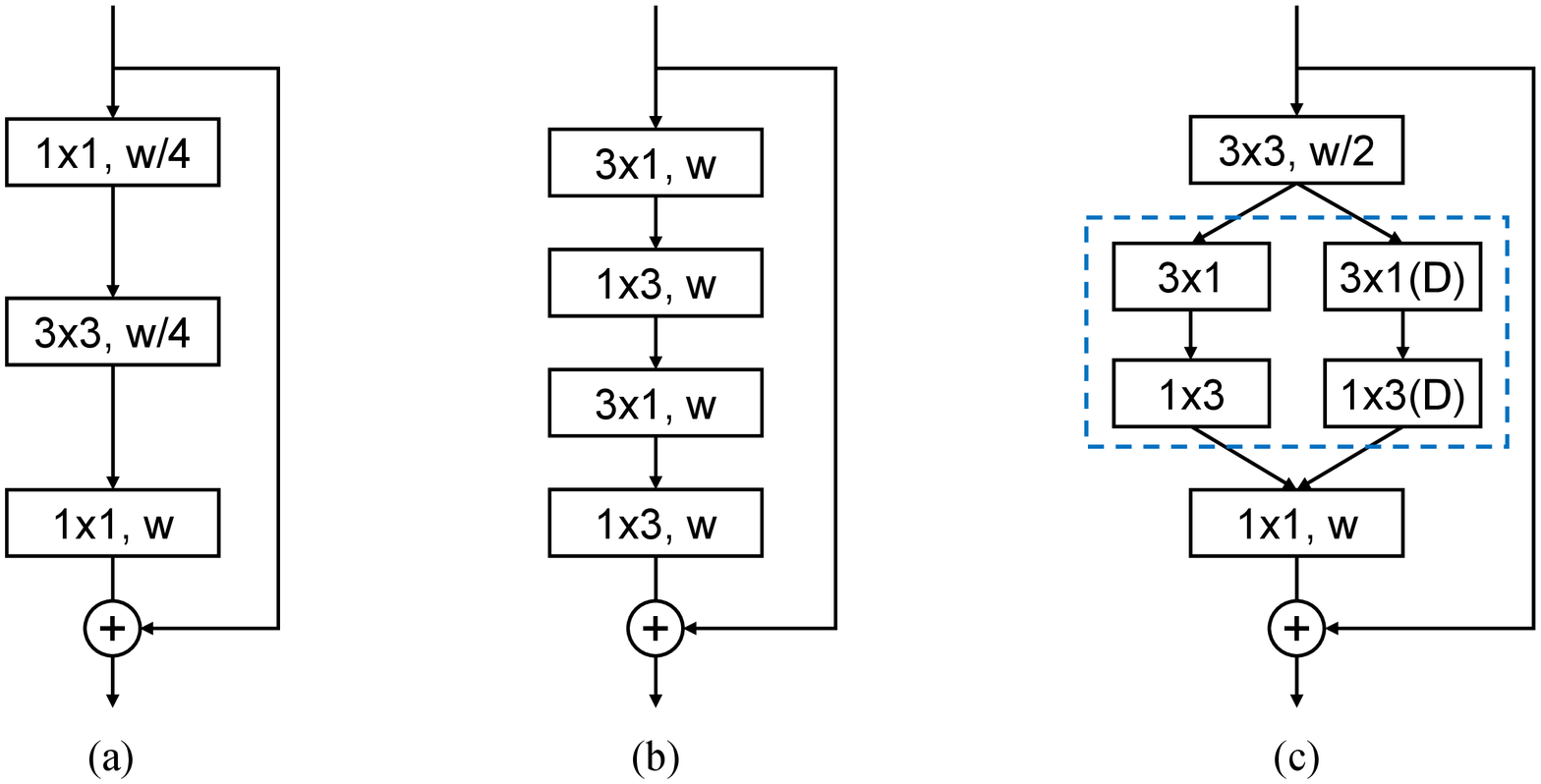}
\caption{(a) ResNet bottleneck design. (b) ERFNet non-bottlenck-1D module. (c) Our DAB module (Convolutions in the dashed box are depth-wise convolution). ``W'': The number of input channels. ``D'': Dilated convolution. For brevity, we omit normalization and activation function.}
\label{Figure2}
\end{figure}
\section{Proposed Network}
In this section, we first introduce the DAB module, which is the core component of DABNet.
Then based on DAB module, we elaborate design choices of the final model, the overall structure of proposed DABNet is shown in Figure \ref{Figure1}.

\subsection{Depth-wise Asymmetric Bottleneck}
\label{Sec3.1}
Inspired by bottleneck design in ResNet \cite{ResNetv1} (See Figure \ref{Figure2}(a)) and factorized convolutions in ERFNet \cite{ERFNet} (See Figure \ref{Figure2}(b)), we design the DAB module with their common advantages.

\noindent\textbf{Bottleneck design.} 
DAB module follows the bottleneck structure similar to ResNet \cite{ResNetv1}.
In order to reduce the parameters and accelerate training process, ResNet-50, 101, 152 modify the original block as a bottleneck design, which is shown in Figure \ref{Figure2}(a).

For the same reason, here we also utilize the bottleneck structure in DAB module, as shown in Figure \ref{Figure2}(c).
Every bottleneck firstly reduces the number of channels by half and restores original channels by a point-wise convolution.
Nevertheless, we make some modifications on the original bottleneck structure:
(1) We use $3\times 3$ convolution at the beginning of each DAB module.
Here we clarify the reason: although a $1\times 1$ convolution has fewer parameters than $3\times 3$ convolution, the intention of ResNet \cite{ResNetv1} is to make a deep model with more than 100 layers, since deep convolutional neural network can increase the receptive field and capture more complex features.
Unfortunately, higher number of layers also bring increasing runtime and memory requirements.
Therefore, for the sake of high inference speed, we instead use the $3\times 3$ convolution to avoid making a deep model.
(2) We only reduce the number of channels by half after the first convolution, as the maximum number of channels in our model is only 128 compared with thousands of channels in ResNet \cite{ResNetv1}.
Thus, for preserving spatial information, we do not compress the channels too much.

\noindent\textbf{Two-branch structure.}
For semantic segmentation, previous remarkable networks have already demonstrated the significance of semantic context information, good prediction results usually come from a combination of hierarchical information, consequently how to effectively fuse multi-scale information remains a tough problem.
In DAB module,  we design a novel two-branch structure and each branch is responsible for extracting corresponding information.

To extract local information, we use a simple $3\times 3$ depth-wise convolution in the first branch. For further reducing the computation, referred by ERFNet \cite{ERFNet} non-bottleneck-1D module (Figure \ref{Figure2}(b)), we apply convolution factorization to depth-wise convolution.
Namely, a standard $n\times n$ depth-wise convolution is substituted for an $n\times 1$ depth-wise convolution followed by a $1\times n$ depth-wise convolution. For an $N\times N$ kernel, asymmetric convolution reduces the computational complexity per pixel from $O(N^2)$ to $O(N)$.  

To extract broader context information, we adopt dilated convolution, which creates a large receptive field without decreasing the resolution of the feature map.
However, when dilation rate becomes larger and larger (in most cases, it increases up to 16 or more), we need to implement lots of padding to maintain the size of the feature map.
This brings heavy computational cost, which prohibits dilated convolution from being used in a real-time model.
In order to resolve the above contradiction, the second branch only applies dilated convolution to the depth-wise asymmetric convolution to reduce computational cost, thus it is called ``depth-wise asymmetric dilated convolution''.

In summary, all the convolutions we used in the ``neck'' part of our bottleneck structure are depth-wise convolution.
In the two-branch structure, each branch can be termed as local information branch and context information branch on the basis of their respective function.
Finally, we add the two branches together and then a $1\times 1$ point-wise convolution is employed at the end of each DAB module, which is used to restore the number of channels and fuse all the channel information.

\noindent\textbf{Activation function.}
In our DAB module, we adopt pre-activation scheme \cite{ResNetv2} and batch normalization \cite{Batchnormal} is used before every non-linear function.
Referring to ENet \cite{ENet}, we use PReLU \cite{PReLU} as nonlinearity function, as PReLU achieves slightly better performance than ReLU due to the shallow network model.
And as mentioned in \cite{MobileNetv2}, using non-linear layers in bottlenecks hurts the performance, therefore non-linearity is removed after the final $1\times 1$ point-wise convolution.

\begin{table}[tbp] \footnotesize
\centering
\begin{tabular}{@{}clccc@{}} 
\toprule
\textbf{Layer} & \textbf{Operator} & \textbf{Mode} & \textbf{Channel} & \textbf{Output size} \\ \midrule
1     & $3\times3$ Conv  & stride 2    & 32    &  $256\times512$     \\
2     & $3\times3$ Conv & stride 1    & 32    &  $256\times512$     \\
3     & $3\times3$ Conv  & stride 1    & 32    &  $256\times512$     \\ \midrule
4     & Downsampling          & -    & 64    &  $128\times256$     \\
5-7   & $3 \times$ DAB module       & dilated 2    & 64    &  $128\times256$   \\ \midrule
8     & Downsampling    & -    & 128   & $64\times128$        \\
9-10  & $2 \times$ DAB module       & dilated 4    &  128  & $64\times128$     \\
11-12 & $2 \times$ DAB module       & dilated 8    &  128  & $64\times128$     \\
13-14 & $2 \times$ DAB module       & dilated 16   & 128   & $64\times128$     \\ \midrule
15    & $1\times1$ Conv            & stride 1     & 19      & $64\times128$    \\
16    & Bilinear interpolation     & $\times8$            & 19      & $512 \times 1024$ \\ \bottomrule
\end{tabular}
\vspace{0.3cm}
\caption{Detailed structure of proposed DABNet.}
\label{Table1}
\end{table}

\subsection{Network Architecture Design}
\label{Sec3.2}
Based on the DAB module, we elaborately design the architecture of DABNet as shown in Figure \ref{Figure1}.
In this subsection, we discuss the design choices on the final model of DABNet.
Since our research aims at the lightweight model with fast inference speed and comparable results, the essence is to build a shallow model with fewer parameters and remove redundant components.
The detailed architecture of DABNet is presented in Table \ref{Table1}.

\noindent\textbf{Downsampling.}
We first use three $3\times 3$ convolutions to extract initial features, then we employ the same downsampling block with ENet \cite{ENet} initial block, which is a concatenation of a $3\times 3$ convolution with stride 2 and a $2\times 2$ max-pooling.
Similarly, we only use the initial block in the first downsampling, the second downsampling block is a single $3\times 3$ convolution with stride 2.

Downsampling operation reduces the size of the feature map and enlarges the receptive field for extracting more contextual information.
However, a reduction in the resolution of the feature map often results in information loss, which has a severe effect on the final prediction result.
Therefore, to preserve the spatial information and details, we only employ three downsampling operations in our model and finally obtain $1/8$ feature map resolution, while most of existing semantic segmentation models employ five downsampling operations and get $1/32$ feature map resolution.
Furthermore, inspired by ESPNetv2 \cite{ESPNetv2}, we build a long-range shortcut connection between the input image and each downsampling block as well as the last convolution layer (see Figure \ref{Figure1}), which facilitates feature reuse and compensates information loss.
This connection first downsamples the original image to the same size as the feature map and then concatenates them together.

\noindent\textbf{DAB block.}
We design two DAB blocks in the DABNet, which include several consecutive DAB modules for dense feature extraction.
The first and the second DAB block consist of 3 and 6 DAB modules, respectively.

To better strengthen spatial relationships and feature propagation, we introduce inter-block concatenation to combine high-level feature with low-level feature, which means stacking first DAB module with last DAB module in each DAB block.
Besides, we employ dilated convolution in every DAB module as mentioned in Section \ref{Sec3.1}.
Specifically, all the DAB module in DAB block 1 include a depth-wise asymmetric dilated convolution with dilation rate 2, and dilation rates in DAB block 2 are 4, 4, 8, 8, 16, 16, respectively.
As with \cite{Dilation8}, we choose this sequential scheme to enlarge the receptive field gradually.

\noindent\textbf{Design choices.}
Note that DABNet is lack of decoder structure, which is much different from popular encoder-decoder structure utilized in most of the semantic segmentation models.
In pursuit of faster inference speed, we discard the decoder in DABNet.
Although the decoder can lead to an effective increase in accuracy, even a small decoder would slow down the network and bring extra computation.
Besides, as we only downsample the input image three times, decoder is not essential in our network.

To make an end-to-end deep learning architecture, we do not use any post-processing technique to improve the final result.
Also, our model does not depend on any backbone, we design our model from scratch.
It is worth noting that the capacity of DABNet is extremely low, and we employ less than 0.76 million parameters.
\section{Experiments}

In this section, we evaluate our proposed network on two challenging datasets: Cityscapes \cite{Cityscapes} and Camvid \cite{CamVid}, which are widely used in semantic segmentation.
Firstly, we introduce the datasets and implementation protocol. 
Then, we conduct several experiments on Cityscapes validation set to prove the effectiveness of our network.
Finally, we report the comparisons with state-of-the-art systems, all the performances are measured using mean intersection-over-union (mIoU).

\subsection{Experimental Settings}
\textbf{Cityscapes.} 
The Cityscapes is a large urban street scene dataset.
 It contains a train set of 2975 images, a validation set of 500 images and a test set of 1525 images.
There is another set of 19,998 images with coarse annotation, but we only use the fine annotated images for all experiments.
The images have a resolution of $1024\times2048$ and 19 semantic categories.

\noindent\textbf{CamVid.}
The CamVid is another street scene dataset for autonomous driving applications.
It includes 701 images in total, 367 for training, 101 for validation and 233 for testing.
The images have a resolution of $360\times480$ and 11 semantic categories.

\noindent\textbf{Implementation protocol.}
All the experiments are performed with one 1080Ti GPU, CUDA 9.0 and cuDNN V7 on the Pytorch platform.
Runtime evaluation is performed on a single 1080Ti card, we report an average of 100 frames for the frames per second (FPS) measurement.

We use mini-batch stochastic gradient descent (SGD) with batch size 8, momentum 0.9 and weight decay $1e^{-4}$ in training.
We employ the ``poly'' learning rate policy \cite{DeepLabv2} and the initial learning rate is set to $4.5e^{-2}$ with power 0.9.
As no pre-training is used, here we set the maximum number of epochs to 1000.
For data augmentation, we employ random horizontal flip, the mean subtraction, and random scale on the input images during training.
The random scale contains \{0.75, 1.0, 1.25, 1.5, 1.75, 2.0\}. Finally, we randomly crop the image into fixed size for training.

\subsection{Ablation Study}
In this subsection, we design a series of experiments to demonstrate the effectiveness of our network.
We adopt Cityscapes dataset \cite{Cityscapes} to conduct quantitative and qualitative analysis.
All the ablation studies are trained from Cityscapes training set and evaluated on Cityscapes validation set.

\begin{table}[tbp]  \footnotesize
\centering
\begin{tabular}{@{}lccc@{}}
\toprule
\textbf{Model} & \textbf{mIoU (\%)} & \textbf{FPS} & \textbf{Parameters (M)} \\ \midrule
DABNet-baseline   & 69.1                  &  104.2            & 0.76                       \\ \midrule\midrule
\multicolumn{4}{l}{(a) Accuracy} \\ \midrule
DABNet-(r=4)   & 66.8   & 104.3   & 0.76           \\ 
DABNet-(r=3, 3, 7, 7, 13, 13) & 68.4   & 104.2             & 0.76                       \\ 
DABNet-ERFdecoder & 69.4    & 58.6            & 1.02                       \\ 
DABNet-SPP  &  68.6                 & 72.2             & 3.22                        \\ \midrule\midrule
\multicolumn{4}{l}{(b) Speed} \\ \midrule
DABNet-w/o dilation &    -               & 104.5             & 0.76                       \\ 
DABNet-First 3x3 conv(r=2) &    -               &  85.6            & 0.76                       \\ 
\bottomrule
\end{tabular}
\vspace{0.3cm}
\caption{Ablation study results on Cityscapes validation set. FPS are estimated for an input of $512 \times 1024$ on a single GTX 1080Ti.}
\label{Table2}
\end{table}

\noindent\textbf{Dilation rates.}
We adopt gradually increasing sequence of dilation rates in the DAB block 2, which is \{4, 4, 8, 8, 16, 16\}.
To investigate the effectiveness of this scheme, we set all the dilation rate as 4 in the DAB block 2 for comparison.
In addition, previous work \cite{UnderstandingCF}  proposed that coprime dilation rates yield better results, therefore we set another varying dilation rates list, which is \{3, 3, 7, 7, 13, 13\}.
In Table \ref{Table2}(a), DABNet-(r=4) performs $2.3\%$ lower accuracy than DABNet, which indicates the importance of increasing dilation rates.
Also, coprime dilation rates perform worse than DABNet, it seems large dilation rates are more effective in our network.

\noindent\textbf{Decoder.}
For striking a balance between the speed and prediction performance, we discard the decoder module in DABNet.
Here we build a network with the decoder of ERFNet for comparison and use deconvolution layer to restore the resolution of the feature map.
As can be seen from Table \ref{Table2}(a), adding the ERF decoder can bring $0.3\%$ slightly better accuracy, while resulting in increased runtime with only $58.6$ FPS.
 Apparently, the decoder is not essential in DABNet, which heavily slows down the model.

\noindent\textbf{Context module.}
PSPNet \cite{PSPNet} is a classic model which employs a spatial pyramid pooling (SPP) module to improve performance by capturing global and local context at different feature resolutions.
To explore the ability to capture context information, we construct a DABNet variant with an SPP head in the end, which is termed as ``DABNet-SPP''.
Table \ref{Table2}(a) shows that DABNet-SPP even decreases $0.5\%$ accuracy, also it has $4.2\%$ times more parameters and slows down the model by 32 FPS.
Therefore, we conclude that the DAB module can better capture context information than a heavy SPP context module.

\noindent\textbf{Inference speed.}
As mentioned in Section \ref{Sec3.1}, dilated convolution brings heavy computational cost and slows down the model.
In order to explore the effect of dilated convolution on inference speed, we design two experiments for comparison in FPS.
In the first experiment we remove all the dilated convolutions in DABNet, and in the other experiment we further adopt a dilation rate 2 in the first $3\times 3$ standard convolution of our DAB module.
As presented in both Table \ref{Table2} (a) and (b), even if we decrease the dilation rate or remove all the dilated convolutions, the FPS is hardly changed (range from 104.2 to 104.5).
Nevertheless, when we apply dilated convolution (with only a dilation rate 2) to the standard convolution, there is an obvious decrease in speed, from 104.2 to 85.6. 
The experimental result proves that dilated convolution has a significant effect on inference speed, but when applied to depth-wise convolution, it has nearly no harmful effect.

\begin{table}[t]\footnotesize 
\centering
\begin{tabular}{@{}lcccccc@{}}
\toprule
Method        & Pretrain &InputSize  & mIoU (\%)     & FPS    & GPU       & Parameters (M) \\ \midrule
DeepLab-v2 \cite{DeepLabv2}    & ImageNet \cite{ImageNet}  & $512\times1024$ & 70.4          &  <1           & TitanX   & 44             \\
PSPNet \cite{PSPNet} & ImageNet & $713\times713$ & 78.4          &<1 & TitanX   & 65.7              \\ \midrule \midrule
SegNet \cite{SegNet} & ImageNet & $360\times640 $ & 56.1     & 14.6    & TitanX  & 29.5           \\
ENet \cite{ENet} & No   & $512\times1024 $     & 58.3          & 76.9            & TitanX  & \textbf{0.36}  \\
SQ \cite{SQ}           & ImageNet  & $1024\times2048 $  & 59.8     & 16.7             & TitanX    & -              \\
ESPNet \cite{ESPNet}       & No    & $512\times1024 $     & 60.3          & \textbf{112}      & TitanX-P    & \textbf{0.36}  \\
ContextNet \cite{ContextNet}    & No    & $1024\times2048 $     & 66.1        & 18.3      & TitanX   & 0.85           \\
ERFNet  \cite{ERFNet}      & No & $512\times1024 $        & 68.0        & 41.7            & TitanX  & 2.1            \\
BiSeNet \cite{BiSeNet}      & ImageNet & $768\times 1536 $  & 68.4  & 105.8            & TitanXp   & 5.8            \\
ICNet  \cite{ICNet}       & ImageNet & $1024\times 2048 $ & 69.5  & 30.3    & TitanX & 7.8              \\ \midrule
DABNet (Ours) & No     & $1024\times2048 $     & \textbf{70.1}      & 27.7      & 1080Ti    & 0.76           \\ \bottomrule
\end{tabular}
\vspace{0.3cm}
\caption{Evaluation results on the Cityscapes test set. TitanX represents the TitanX Maxwell, TitanX-P represents the TitanX Pascal. (The GPU computational efficiency: TitanX Maxwell <\ Titan X Pascal $\approx$ 1080Ti < TitanXp)}
\label{Table3}
\end{table}

\begin{table}[t] \footnotesize
\centering
\begin{tabular}{@{}lcccc@{}}
\toprule
Method        & mIoU (\%)  & Parameters (M) \\ \midrule 
ENet \cite{ENet}        & 51.3  & \textbf{0.36}  \\
SegNet \cite{SegNet}  & 55.6    & 29.5           \\
FCN-8s \cite{FCN}       & 57.0   & 134.5          \\
Dilation8 \cite{Dilation8}    & 65.3    & 140.8          \\
BiSeNet \cite{BiSeNet}     & 65.6  & 5.8            \\
ICNet \cite{ICNet}   & \textbf{67.1}     & 26.5             \\ \midrule	
DABNet(Ours)  & 66.4       & 0.76
\\ \bottomrule
\end{tabular}
\vspace{0.3cm}
\caption{Evaluation results on the CamVid test set.}
\label{Table4}
\end{table}

\begin{figure}[htb]
\centering
\includegraphics[scale=0.23]{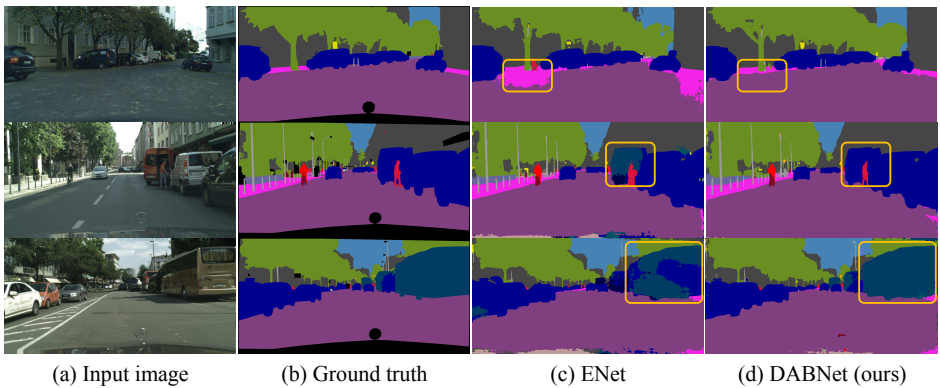}
\caption{Qualitative examples of the segmentation on Cityscapes validation set. From left to right: Input image, ground-truth, prediction of ENet, and prediction of DABNet.}
\label{Figure3}
\end{figure}

\begin{table}[t] \footnotesize
\centering
\begin{tabular}{@{}lcccccc@{}}
\toprule
\multirow{3}{*}{Model} & \multicolumn{6}{c}{GTX 1080Ti}                                                                               \\ \cmidrule(l){2-7} 
                       & \multicolumn{2}{c}{$256\times512$}                      & \multicolumn{2}{c}{$512\times1024$} & \multicolumn{2}{c}{$1024\times2048$} \\
                       & \multicolumn{1}{c}{ms} & \multicolumn{1}{c}{fps} & \multicolumn{1}{c}{ms} & \multicolumn{1}{c}{fps} & \multicolumn{1}{c}{ms} & \multicolumn{1}{c}{\ \ \ fps}           \\ \midrule
SegNet \cite{SegNet}     & 16                & 64.2                         &  56                      & 17.9    &  -             &  -             \\
ENet \cite{ENet}       &    10                    &  99.8                       &       13                 & 74.9    &    44           &22.9               \\
ICNet \cite{ICNet}       &    9                    &  107.9                       &       15                 & 67.2    &    40           &25.1            \\
ESPNet \cite{ESPNet} &5 &182.5 & 9 & 115.2 & 30 & 33.3 \\ \midrule
DABNet (Ours)           & 6                        &  170.2                       &   10                     & 104.2    &         36      &  27.7             \\ \bottomrule
\end{tabular}
\vspace{0.3cm}
\caption{Speed comparison of our method against other state-of-the-art methods. All the runtime is computed on a single GTX 1080Ti. (``-'' indicates the model exceeds the maximum memory of a single GTX 1080Ti.)}
\label{Table5}
\end{table}

\subsection{Comparison with state-of-the-arts}
In this subsection, we compare our algorithm with state-of-the-art models.
We first report the final results on Cityscapes \cite{Cityscapes} and CamVid \cite{CamVid} benchmarks,
then we analyze the speed of our model and compute the FPS of other state-of-the-art methods under the same status for fair comparison.
During evaluation, we do not adopt any testing tricks such as multi-crop and multi-scale testing.

\noindent\textbf{Accuracy and parameter comparisons.}
With only 0.76 million parameters, our DABNet achieves impressive results.
Here we show the results on Cityscapes and CamVid benchmarks, respectively.
As shown in Table \ref{Table3}, DABNet achieves 70.1\% mIoU on Cityscapes test set, which significantly outperforms existing real-time segmentation work.

In general, the more parameters, the more complex the function that can be fitted.
However, most of the large models have massive redundant parameters.  
Table \ref{Table3} shows that DABNet only uses 9.7\% of ICNet's parameters, but achieves a better result.
Moreover, as can be observed, compared with the offline methods---PSPNet \cite{PSPNet} and DeepLabV2 \cite{DeepLabv2}, we only use 1\% of their parameters to keep a comparable performance.
The qualitative results on Cityscapes validation set are presented in Figure \ref{Figure3}.
On CamVid test set, as reported in Table \ref{Table4}, DABNet again achieves outstanding performance with small capacity, and it can process a $360\times480$ CamVid image at the speed of 146 FPS.

\noindent\textbf{Speed comparison.} 
Since different methods employ various devices and diverse input sizes, 
for fair comparison of speed, all the speed experiments are performed with a single 1080Ti GPU on the Pytorch platform.
Additionally, we also list the FPS and their test device from existing literature in Table \ref{Table3} for reference.
Table \ref{Table5} presents the speed comparison between our method with other state-of-the-art approaches at full, half and quarter resolution on images of Cityscapes.
DABNet is able to process a $1024\times 2048$ image at the speed of 27.7 FPS.
ESPNet \cite{ESPNet} is one of the fastest real-time networks, and it has slightly quicker speed than DABNet. However it only achieves 60.3\% mIoU which is 9.8\% less than our network.
The speed comparison represents that DABNet is able to process a high resolution image with fast inference speed while maintaining high performance.

\section{Conclusions}
In this paper, we propose a novel Depth-wise Asymmetric Bottleneck module to extract local and contextual features jointly.
Based on the DAB module, we elaborately design the Depth-wise Asymmetric Bottleneck Network, a lightweight model with fast inference speed and competitive results. 
Through analysis and quantitative experimental results, we demonstrate the effectiveness of our method.
The proposed DABNet achieves 70.1\% Mean IoU on Cityscapes test set with only 0.76 million parameters, and can run at 104 fps on $512\times 1024$ high-resolution images.
In conclusion, our network shows significant improvements in all the aspects of accuracy, speed, and capacity compared with the state-of-the-art methods.

\section*{Acknowledgement}
This work was supported by the National Research Foundation of the Korea (No.NRF-2017R1D1A1B03034288)
and the MSIT (Ministry of Science, ICT), Korea, under the High-Potential Individuals Global Training Program (2019-0-01609) supervised by the IITP (Institute for Information \& Communications Technology Planning \& Evaluation).

\bibliography{egbib}
\end{document}